\documentclass[10pt,twocolumn,letterpaper]{article}

\usepackage{cvpr}
\usepackage{times}
\usepackage{epsfig}
\usepackage{graphicx}
\usepackage{amsmath}
\usepackage{amssymb}
\usepackage{dsfont}
\usepackage{arydshln}
\usepackage{bm}
\usepackage{tabto}
\usepackage{subcaption}
\usepackage{booktabs}
\usepackage{tabularx} 
\usepackage{enumitem}
\usepackage{textcomp}
\DeclareMathOperator*{\argmax}{arg\,max}
\newenvironment{tight_enumerate}{
\begin{enumerate}
  \setlength{\itemsep}{0pt}
  \setlength{\parskip}{0pt}
}{\end{enumerate}}

\newenvironment{nscenter}
 {\parskip=0pt\par\nopagebreak\centering}
 {\par\noindent\ignorespacesafterend}

\usepackage[breaklinks=true,bookmarks=false]{hyperref}
\cvprfinalcopy

\author{ {Thomas Verelst \qquad  Tinne Tuytelaars}  \\ 
{ESAT-PSI, KU Leuven} \\ 
\tt\small \{thomas.verelst, tinne.tuytelaars\}@esat.kuleuven.be}

\def\cvprPaperID{6535} 

\ifcvprfinal\fi

 \AddToShipoutPicture*{\scriptsize \sffamily\raisebox{1.5cm}{\hspace{2.4cm}
\begin{minipage}{16cm}
DOI:  10.1109/CVPR42600.2020.00239 \quad \texttt{https://ieeexplore.ieee.org/abstract/document/9157150} \quad © 2020 IEEE. Personal use of this material is permitted. Permission from IEEE must be
obtained for all other uses, in any current or future media, including
reprinting/republishing this material for advertising or promotional purposes, creating new
collective works, for resale or redistribution to servers or lists, or reuse of any copyrighted
component of this work in other works.
\end{minipage}
 }}

\begin{document}

\title{Dynamic Convolutions: Exploiting Spatial Sparsity for Faster Inference}

\maketitle

\begin{abstract}
Modern convolutional neural networks apply the same operations on every pixel in an image.
However, not all image regions are equally important. To address this \mbox{inefficiency}, we propose a method to dynamically apply convolutions conditioned on the input image. We introduce a residual block where a small gating branch learns which spatial positions should be evaluated. These discrete gating decisions are trained end-to-end using the Gumbel-Softmax trick, in combination with a sparsity criterion. Our experiments on CIFAR, ImageNet, Food-101 and MPII show that our method has better focus on the region of interest and better accuracy than existing methods, at a lower computational complexity. Moreover, we provide an efficient CUDA implementation of our dynamic convolutions using a gather-scatter approach, achieving a significant improvement in inference speed on MobileNetV2 and ShuffleNetV2. On human pose estimation, a task that is inherently spatially sparse, the processing speed is increased by 60\% with no loss in accuracy.
\thispagestyle{empty}

\end{abstract}


\section{Introduction}

Most research on deep neural networks focuses on improving accuracy without taking into account the model complexity. 
As the community moves to
more difficult problems -- \eg~from classification to detection or pose estimation -- architectures tend to grow in capacity and computational complexity.
Nevertheless, 
for real-time applications running on consumer devices such as mobile phones, notebooks or surveillance cameras, what matters most is a good trade-off between performance
(i.e., frames processed per second) and accuracy~\cite{canziani2016analysis, kim2015compression}. 
Attempts to improve this trade-off have focused mostly on designing more efficient architectures~\cite{iandola2016squeezenet, mobilenetv2, mnasnet, shufflenet}
or compressing existing ones~\cite{han2015deep,hinton2015distilling,li2016pruning,molchanov2016pruning, sainath2013low, wu2016quantized}. 

Interestingly,
most neural networks, including the more efficient or compressed ones mentioned above, 
execute the same calculations for each image, 
independent of its content. This seems suboptimal: only the complex images require such deep and wide networks. 
Therefore, the domain of conditional execution gained 
momentum~\cite{bengio2015conditional,bengio2013deep, bengioST}. Compared to static compression methods, the architecture of the network is adapted based on the input image. For instance, the network depth can vary per image since easy and clear images require fewer convolutional layers than ambiguous ones~\cite{convnetaig,skipnet,blockdrop}. The neural network chooses which operations to execute. Such practice is often called \textit{gating}~\cite{droniou2015deep, jacobs1991adaptive}, and can be applied at the level of convolutional layers~\cite{ convnetaig,skipnet,blockdrop}, channels~\cite{bejnordi,gao,lin2017runtime} or other elements in the network. 

\begin{figure}[t]
\centering
\includegraphics[width=1\linewidth]{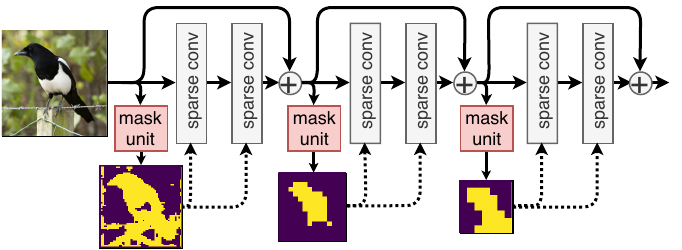}
\caption{In each residual block, a small gating network {(mask unit)} predicts pixel-wise masks determining the locations where dynamic convolutions are evaluated.}
\label{fig:arch_overview}
\end{figure}

In this work, we focus on reducing computations by {\em executing conditionally in the spatial domain}. Typical convolutional networks apply all convolutional filters on every location in the image. 
In many images, the subject we want to classify or 
detect is surrounded by background pixels, where the necessary features can be extracted using only few operations. For example, flat regions such as a blue sky can easily be identified. We call such images spatially sparse.

We propose a method, trained end-to-end without explicit spatial supervision, to execute convolutional filters on important image locations only. 
For each residual block, a small gating network chooses the locations to apply dynamic convolutions on (Fig.~\ref{fig:arch_overview}). Gating decisions are trained end-to-end using the Gumbel-Softmax trick~\cite{gumbelsoftmax,maddison}. 
Those decisions progress throughout the network: the first stages extract features from complex regions in the image, while the last layers use higher-level information to focus on the region of interest only.
Note that the input of a dynamic convolution is a dense matrix, making it fundamentally different from methods operating on sparse data~\cite{graham2014spatially,graham2017submanifold}. 

Many works on conditional execution only report a reduction in the theoretical complexity~\cite{sact,convnetaig}.
When implemented naively,
merely applying masks does not save computations. For certain methods, it is actually not clear whether they could be implemented efficiently at all. For instance, Wu \etal~\cite{blockdrop} report an \emph{increase} in execution time when conditionally executing individual layers using a separate policy network. Parallel execution on GPU or FPGA relies on the regularity of standard convolutions to pipeline operations~\cite{lavin2016fast,sze2017efficient} and adding element-wise conditional statements might strongly slow down inference.  
    Moreover, Ma \etal~\cite{shufflenet} show that the number of floating point operations (FLOPS) is not a sufficient metric to estimate inference speed: networks with a similar number of operations can have largely different execution speeds. 
    Simple element-wise operations such as activation functions, summations and pooling can have a significant impact, but are not included in many FLOPS-estimations. 

In contrast, we demonstrate an actual improvement of wall-clock time with our CUDA implementation of dynamic convolutions. Our method is designed with practical feasibility in mind and requires minimal changes to existing libraries: efficient spatially sparse execution is made possible by rearranging tensor elements in memory, which has similar overhead as a simple copy operation. Our code is available at { \texttt{https://github.com/thomasverelst/dynconv}}.

The main contributions of our paper are threefold:
\begin{tight_enumerate}
    \item We present an approach to train pixel-wise gating masks end-to-end using the Gumbel-Softmax trick, with a focus on efficiency. 
    \item Our method achieves state-of-the-art results on classification tasks with ResNet~\cite{resnet} and MobileNetV2~\cite{mobilenetv2}, and we show strong results on human pose estimation, improving the performance-accuracy trade-off over non-conditional networks.
    \item We provide a CUDA implementation of residual blocks with dynamic convolutions on GPU, not just reducing the theoretical number of floating-point operations but also offering practical speedup with MobileNetV2 and ShuffleNetV2.
\end{tight_enumerate}

\section{Related work}

Static compression methods have been extensively studied to reduce storage and computational cost of existing neural networks, \eg pruning~\cite{lecun1990optimal,li2016pruning, molchanov2016pruning}, knowledge distillation~\cite{hinton2015distilling, romero2014fitnets}, structured matrices~\cite{sindhwani2015structured,wen2016learning} or quantization~\cite{han2015deep, wu2016quantized}. Recent methods vary computations
based 
on the input image. So-called conditional execution can be applied on several aspects of a network: we make a distinction between 
layer-based, channel-based and spatial methods.

\textbf{Layer-based methods} conditionally execute certain network layers or blocks depending on the input. Easy images require a less deep network than complex examples. One of the first methods, Adaptive Computation Time~\cite{graves}, interprets residual blocks as a refinement of features. Execution of layers is halted when features are `good enough' for the classifier.
Another approach is to use early-exit branches in the network~\cite{adaptivenetworks,panda2016conditional,branchynet}.
Later methods improve the flexibility by conditionally executing individual layers. Those methods are based on the observation that residual architectures are robust against layer dropout~\cite{stochasticdepth,veit2016residual}.
\mbox{SkipNet}~\cite{skipnet} learns gating decisions using reinforcement learning. ConvNet-AIG~\cite{convnetaig}  uses the Gumbel-Softmax trick and BlockDrop~\cite{blockdrop} trains a separate policy network using reinforcement learning.

\textbf{Channel-based methods} prune channels dynamically and on-the-fly during inference. The main motivation is that advanced features are only needed for a subset of the images: features of animals might differ from those of airplanes. Gao~\etal~\cite{gao} rank channels and only execute the top-k ones. Lin \etal~\cite{lin2017runtime} propose a method to train an agent for channel-wise pruning using reinforcement learning, while Bejnordi \etal~\cite{bejnordi} use the Gumbel-Softmax trick to gate channels conditionally on the input.


\textbf{Spatial methods} exploit the fact that not all regions in the image are equally important. A first set of methods~\cite{dyanmiccapacity,mnih2014recurrent,sharma2015action} uses glimpses to only process regions of interest. Such a two-stage approach is limited to applications where the object of interest is clearly separated, since all information outside the crop is lost. 
The glimpse idea has been extended for pixel-wise labeling tasks, such as semantic segmentation, using cascades~\cite{li2017cascade}.
Later methods offer a finer granularity and more control over the conditional execution. The closest work to ours is 
probably
Spatially Adaptive Computation Time (SACT)~\cite{sact}. It is a spatial extension of a work by Graves~\cite{graves} and varies the number of residual blocks executed per spatial location. Features are processed until a halting score determines that the features are good enough.
Since the method relies on refinement of features, it is only applicable to residual networks with many consecutive layers. Our method is more general and flexible as it makes skipping decisions per residual block and per spatial location. In addition, they do not show any inference speedup.

One of the only works showing practical speedup with spatially conditional execution is SBNet~\cite{sbnet}. Images are divided into smaller \textit{tiles}, which can be processed with dense convolutions.
A low-resolution network first extracts a mask, and the second main network processes tiles according to this mask. Tile edges overlap to avoid discontinuities in the output, causing significant overhead when tiles are small. Therefore the tile size typically is $16 \times 16$ pixels, which makes the method only suitable for large images. They demonstrate their method on 3D object detection tasks only. In contrast, our approach integrates mask generation and sparse execution into a single network, while providing finer pixel-wise control and efficient inference.

\section{Method}
For each individual residual block, a small gating network generates execution masks based on the input of that block (see Fig.~\ref{fig:arch_overview}). 
We first describe how pixel-wise masks are learned using the Gumbel-Softmax trick. Afterwards, we elaborate on the implementation of dynamic convolutions, used to reduce inference time.
Finally, we propose a sparsity criterion that is added to the task loss in order to achieve the desired reduction in computations.


\subsection{Trainable masks}

Pixel-wise masks define the spatial positions to be processed by convolutions. These discrete decisions, for every spatial location and every residual block independently, are trained end-to-end using the Gumbel-Softmax trick~\cite{gumbelsoftmax}. 


\subsubsection{Block architecture}

\label{sec:block_arch}
Denote the input of the residual block $b$ as $X_b \in \mathds{R}^{c_b \times w_b \times h_b}$.
Then the operation of the residual block is described by
\begin{equation}
\label{eq:residual}
X_{b+1} = {r}(\mathcal{F}(X_b) + X_b)
\end{equation}
with $\mathcal{F}$ the residual function, typically two or three convolutions with batchnorm (BN), and $r$ an activation function. Our work makes $\mathcal{F}$ conditional in the spatial domain. A small mask unit  $\mathcal{M}$ outputs soft gating decisions $M_b \in \mathds{R}^{w_{b+1} \times h_{b+1}}$, based on input $X_b$. 
We compare the mask unit of SACT~\cite{sact} (referred to as \textit{squeeze unit}), incorporating a squeeze operation over the spatial dimensions, to a $1{\times}1$ convolution. We use the squeeze unit in classification, and the $1{\times}1$ convolution for pose estimation.

\begin{figure}[t]
\begin{nscenter}
   \includegraphics[width=1\linewidth]{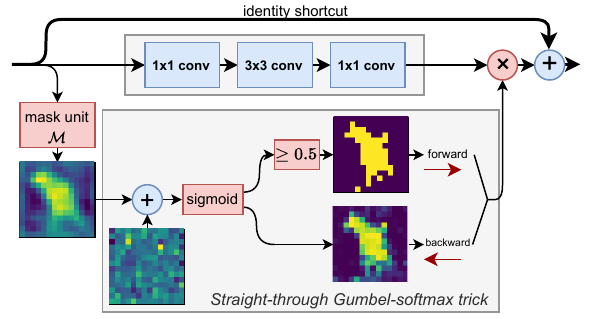}
\end{nscenter}
   \caption{Training spatial execution masks using the Gumbel-Softmax trick. The mask unit generates a floating-point mask, after which the Gumbel-Softmax trick converts soft-decisions into hard-decisions and enables backpropagation for end-to-end learning. }
\label{fig:block_training}
\end{figure}

The Gumbel-Softmax module $\mathcal{G}$ turns soft decisions $M_b$ into hard decisions $G_b  \in \{0,1\}^{w_{b+1} \times h_{b+1}}$ by applying a binary Gumbel-Softmax trick (see Section~\ref{sec:binary_gumbel}) on each element of $M_b$:
\begin{equation}
\label{eq:residual_conditional}
G_{b} =\mathcal{G}(\mathcal{M}(X_b)).
\end{equation}
Gating decisions $G_b$ serve as \emph{execution masks}, indicating active spatial positions where the residual block should be evaluated.
The residual block with spatially sparse inference is then described by
\begin{equation}
\label{eq:residual_conditioanl}
X_{b+1} = {r}(\mathcal{F}(X_b)\circ G_b + X_b) 
\end{equation}
with $\circ$ an element-wise multiplication over the spatial dimensions  $(w_{b+1} \times h_{b+1})$, broadcasted over all channels. During training, this is an actual multiplication with the mask elements in order to learn gating decisions end-to-end (see Fig.~\ref{fig:block_training}). During inference, the residual function is only evaluated on locations indicated by the execution mask $G_b$.

\subsubsection{Binary Gumbel-Softmax}
\label{sec:binary_gumbel}
The Gumbel-Softmax trick turns soft decisions into hard decisions while enabling backpropagation, needed to optimize the weights of the mask unit. Take a categorical distribution with class probabilities $\bm{\pi}~=~{\pi_1, \pi_2,... \pi_n}$, then discrete samples $z$ can be drawn using
\begin{equation}
\label{eq:gumbel_hard}
z = \mathtt{one\_hot}\bigl(  \argmax_i{ [log(\pi_i) + g_i] } \bigr)
\end{equation}
with $g_i$ being noise samples drawn from a Gumbel distribution. 
The Gumbel-Softmax trick defines a continuous, differentiable approximation by replacing the argmax operation with a softmax: 
\begin{equation}
\label{eq:gumbel}
y_i =  \frac {\exp( (log(\pi_i) + g_i ) / \tau)}{ \sum_{j=1}^k  \exp( (log(\pi_j) + g_j ) / \tau)  }.
\end{equation}

Gating decisions are binary, which makes it possible to strongly simplify the Gumbel-Softmax formulation.  A soft-decision $m \in (-\infty, \infty)$, outputted by a neural network, is converted to a probability $\pi_{1}$ indicating the probability that a pixel should be executed, using a sigmoid $\sigma$.  
\begin{align}
\label{eq:probtaken}
\pi_{1} &= \sigma(m).
\end{align}
Then, the probability 
that a pixel is not executed is
\begin{equation}
\label{eq:probnottaken}
\pi_{2} = 1 - \sigma(m) .
\end{equation}

Substituting $\pi_1$ and $\pi_2$ in Equation~\ref{eq:gumbel}, for the binary case of $k=2$ and $i=1$, makes it possible (see supplementary) to reduce this to
\begin{equation}
y_1 = \sigma \Bigl( \frac{ m + g_{1} - g_{2} }{\tau} \Bigr).
\end{equation}


Our experiments use a fixed temperature $\tau = 1$, unless mentioned otherwise. We use a straight-through estimator, where hard samples are used during the forward pass and gradients are obtained from soft samples during the backward pass:
\begin{equation}
 \label{eq:straight_through}
z = 
\begin{cases}
y_1 > 0.5 \equiv \frac{m + g_{1} - g_{2}}\tau > 0&\text{(forward)}, \\
y_1 &\text{(backward)}.
\end{cases}
\end{equation}

Note that this formulation has no logarithms or exponentials in the forward pass, typically expensive computations on hardware platforms. During inference, we do not add Gumbel noise and therefore models are finetuned  
during the last 20 percent of epochs 
with $g_1$ and $g_2$ fixed to 0, making it similar to the straight-through estimator of Bengio \etal~\cite{bengioST}.


\subsection{Efficient inference implementation}
\label{sec:eff_inference}


The residual function should be evaluated on active spatial positions only. Efficiently executing conditional operations can be challenging: hardware strongly relies on regularity to pipeline operations. Especially spatial operations, \eg $3{\times}3$ convolutions, require careful optimization and data caching 
~\cite{lavin2016fast}.


Our method copies elements at
selected spatial positions to an intermediate, dense tensor using a \emph{gather} operation. Non-spatial operations, such as pointwise $1{\times}1$ convolutions and activation functions, can be executed efficiently by applying existing implementations on the intermediate tensor. The result is copied back to its original position afterwards using a \emph{scatter} operation.
More specifically, let the input $I$ of a residual block be a 4D tensor with dimensions $N{\times}C{\times}H{\times}W$, being the batch size, number of channels, height and width respectively. The gather operation copies the active spatial positions to a new intermediate tensor $T$ with dimensions $P{\times}C{\times}1{\times}1$, where $P$ is the number of active spatial positions spread over the $N$ inputs of the batch. The intermediate tensor can be seen as $P$ images of size $1{\times}1$ with $C$ channels, and non-spatial operations can be applied as usual. 
The inverted residual block of MobileNetV2 relies heavily on non-spatial operations and was designed for low computational complexity, making it a good fit for conditional execution. It consists of a pointwise convolution expanding the feature space, followed by a lightweight depthwise (DW) convolution and linear pointwise bottleneck. 
The gather operation is applied before the first pointwise convolution, which then operates on the intermediate tensor $T$. The depthwise convolution is the only spatial operation in the block and should be adapted to operate on the atypical dimensions of $T$. The second pointwise convolution is followed by the scatter operation, after which the residual summation is made.  The architecture of the residual block with dynamic convolutions is shown in Figure~\ref{fig:arch-inference} and next we describe the role of
each component:
\begin{figure}[tb]
\begin{nscenter}
  \includegraphics[width=1\linewidth]{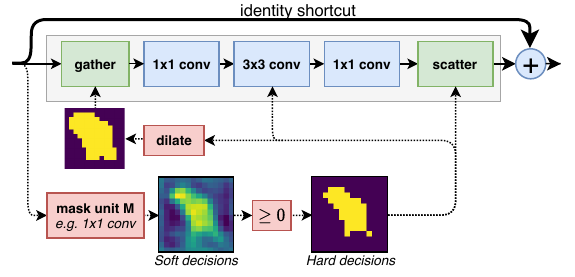}
\end{nscenter}
  \caption{Architecture of a residual block for efficient sparse inference. The mask unit $\mathcal{M}$ generates a mask based on the block's input. The gather operation uses the mask to copy selected spatial positions (yellow) to a new intermediate tensor. Non-spatial operations 
  use standard implementations, while the $3 {\times}3$ convolution is modified to operate on the intermediate tensor.
  }
\label{fig:arch-inference}
\end{figure}
\begin{itemize}[leftmargin=*, topsep=0pt]
    \setlength\itemsep{0em}
     \item \textbf{Mask dilation}:  Gating decisions $G_b$ 
     indicate positions where the spatial $3{\times}3$ convolution should be applied. The first $1{\times}1$ convolution should also be applied on neighboring spatial positions to avoid gaps in the input of the $3{\times}3$ convolution. Therefore, the mask $G_b$ is morphologically dilated, resulting in a new mask $G_{b,dilated}$.
    \item \textbf{Masked gather operation:} The gather operation copies active spatial positions from input $I$ with dimension ${N{\times}{C}{\times}{H}{\times}{W}}$ to an intermediate tensor $T$ with dimension ${P{\times}C{\times}1{\times}1}$. The index mapping from $I$ to $T$ depends on the execution mask $G$: an element $I_{n,c,h,w}$ being the $p$th active position in a flattened version of $G$, is copied to $T_{p,c,1,1}$.
    \item \textbf{Modified 3x3 Depthwise Convolution:} The depthwise convolution applies a $3{\times}3$ convolutional kernel to each channel separately. 
    We implement a custom CUDA kernel that applies the channelwise filtering efficiently on~$T$. The spatial relation between elements of~$T$ is lost due to its dimensions being $P{\times}C{\times}1{\times}1$. When processing elements~$t$ in~$T$, our implementation retrieves the memory locations of spatial neighbors using an index mapping from $T$~to~$I$.
\end{itemize}



Traditional libraries for sparse matrix operations have considerable overhead due to indexing and bookkeeping. Our solution minimizes this overhead by gathering elements in smaller, dense matrix. The extra mapping step in the modified $3{\times}3$ DW convolution has minimal impact since its computational cost is small compared to $1{\times}1$ convolutions. Note that the gather-scatter strategy combined with a modified depthwise convolution can be applied on other networks, such as ShuffleNetV2~\cite{shufflenet} and {MnasNet}~\cite{mnasnet}. 

\subsection{Sparsity loss}
\label{sec:losses}
Without additional constraints, the most optimal gating state is to execute every spatial location. We define a \emph{computational budget} hyperparameter $\theta \in [0,1]$, indicating the relative number of desired operations. For instance, 0.7 means that on average 70\% of the FLOPS in the conditional layers should be executed. The total number of floating point operations for convolutions in a MobileNetV2 residual block $b$ with stride 1 is 
\begin{equation}
\mathds{F}_b =H \cdot W \cdot \big( 9 C_{b,e} + 2 C_b C_{b,e}\big) ,
\end{equation}
with $C_b$ the number of base channels and $C_{b,e}$ the number of channels for the depthwise convolution (typically $6C_b$) and $H{\times}W$ the spatial dimensions.
For sparse inference with dynamic convolutions, this becomes
\begin{equation}
\mathds{F}_{b,sp} = N_{b,dilated}  C_b C_{b,e}  +  N_{b}  \big(9 C_{b,e}  + C_{b,e} C_b \big),
\end{equation}
with $N_{b,dilated}$ and $N_{b}$ the number of active spatial positions in the dilated mask and mask respectively. The value $N_{b}$ is made differentiable by calculating it as the sum of all elements in the execution mask of that block (Eq.~\ref{eq:residual_conditional}):
\begin{align}
N_{b} = \sum{G_b}.
\end{align}


The following loss then minimizes the difference between the given computational budget $\theta$ and the budget used by a network consisting of $B$ residual blocks:
\begin{equation}
\label{eq:loss_net}
\mathcal{L}_{sp,net} = \biggl( \frac{ \sum_b^B \mathds{F}_{b,sp} }{ \sum_b^B \mathds{F}_b } - \theta \biggr)^2 .
\end{equation}
In practice, we average the FLOPS over the batch size, and the network is free to choose the distribution of computations over the residual blocks and batch elements. However, without proper initialization this could lead to a suboptimal state where the network executes all positions in a certain block or none. Minimizing the sparsity loss is trivial compared to the task loss and the mask units never recover from this state. This problem occurs often in conditional execution and existing solutions consist of dense pretraining with careful initialization~\cite{sact,skipnet},
curriculum learning~\cite{blockdrop} or setting a computational budget for each residual block individually~\cite{convnetaig}. The latter can be formulated as
\begin{equation}
\label{eq:loss_layer}
\mathcal{L}_{sp,per\_layer} = \sum_b^B \biggl( \frac{\mathds{F}_{b,sp} }{\mathds{F}_b } - \theta \biggr)^2 .
\end{equation}
Such a constraint per layer is effective but limits the flexibility of computation allocations.
We propose a solution to ensure proper initialization of each block, by adding an extra constraint that keeps the percentage of executed operations ${\mathds{F}_{b,sp}}/{\mathds{F}_b}$ in each residual bock between an upper and lower bound. Those bounds are annealed from the target budget $\theta$ to 0~and~1 respectively. 

The upper and lower bound are imposed by the following loss functions, where we use cosine annealing to vary $p$ from 1 to 0 between the first and last epoch of training:
\begin{align}
&\mathcal{L}_{sp,low} = \frac1B\sum_b^B \max(0, p \cdot \theta - \frac{\mathds{F}_{b,sp}}{\mathds{F}_{b}})^2, \\
&\mathcal{L}_{sp,up} = \frac1B\sum_b^B \max(0, \frac{\mathds{F}_{b,sp}}{\mathds{F}_b}-(1-p(1-\theta)))^2.
\end{align}
The final loss to be optimized is then given by
\begin{equation}
\label{eq:finalloss}
\mathcal{L} = \mathcal{L}_{task} +  \alpha(\mathcal{L}_{sp,net} + \mathcal{L}_{sp,lower} + \mathcal{L}_{sp,upper})
\end{equation}
where $\alpha$ is a hyperparameter, chosen so that the task and sparsity loss have the same order of magnitude. We choose $\alpha = 10$ for classification and $\alpha = 0.01$ for pose estimation. 


\section{Experiments and results}\
We first compare our masking method with other conditional execution approaches on CIFAR and ResNet, and show that our method improves the accuracy-complexity trade-off. Then we demonstrate inference speedup on \mbox{Food-101} with MobileNetV2 and ShuffleNetV2.
Finally we apply our method on pose estimation, a task that is typically spatially sparse. We study the impact of the proposed sparsity criterion on this task. 

\subsection{Classification}





\subsubsection{Comparison with state-of-the-art }


The ResNet~\cite{han2015deep} CNN for classification is typically used to compare the performance of conditional execution methods. We compare the theoretical number of floating point operations and accuracy. 
The main work related to ours is SACT~\cite{sact}, also exploiting spatial sparsity. 
ConvNet-AIG~\cite{convnetaig}, conditionally executing complete residual blocks, can be seen as a non-spatial variant of our method. 

\begin{figure}[tb]
\centering
\begin{subfigure}{.5\linewidth}
  \centering
  \includegraphics[width=1\linewidth]{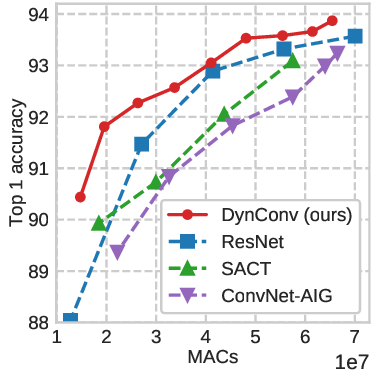}
  \caption{CIFAR-10}
  \label{fig:sota_cifar10}
\end{subfigure}%
\begin{subfigure}{.5\linewidth}
  \centering
  \includegraphics[width=1\linewidth]{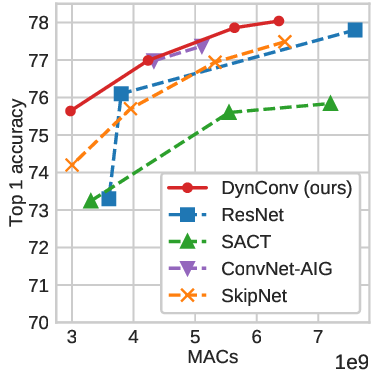}
  \caption{ImageNet}
  \label{fig:sota_imagenet}
\end{subfigure}
\caption{Comparison with state-of-the-art models}
\end{figure}

\paragraph{CIFAR-10} We perform experiments with ResNet-32 on the standard train/validation split of CIFAR-10~\cite{cifar}. We use the same hyperparameters and data augmentation as ConvNet-AIG, being an SGD optimizer with momentum 0.9, weight decay 5e-4, learning rate 0.1 decayed by 0.1 at epoch 150 and 250 with a total of 350 epochs. 
Results for SACT and ConvNet-AIG are obtained with their provided code.  We evaluate our method for different budget targets 
$\theta \in \{0.1,0.2,\dots,0.9\}$. The mask unit is a squeeze unit with the same architecture as in SACT. Non-adaptive baseline ResNet models have $8,14,20,26$ and $32$ layers. Figure~\ref{fig:sota_cifar10} shows that our method (DynConv) outperforms the other conditional execution methods for all computational costs (MACs) while improving the accuracy-complexity trade-off. Moreover, there is a smaller drop in accuracy when reducing the computational budget.  

\paragraph{ImageNet}
We use ResNet-101~\cite{resnet} and ImageNet~\cite{imagenet} to compare DynConv against results reported in SACT~\cite{sact}, ConvNet-AIG~\cite{convnetaig}, SkipNet~\cite{skipnet} and standard ResNet~\cite{resnet}. Just like SACT, we initialize convolutional layers with weights from a pretrained ResNet-101 since the large number of conditional layers makes the network prone to {dead residuals}, where some layers are not used at all. We use the standard training procedure of ResNet~\cite{resnet} with InceptionV3~\cite{szegedy2016rethinking} data augmentation. Models are trained on a single GPU with batch size 64 and learning rate $0.025$ for 100 epochs. Learning rate is decayed by 0.1 at epoch 30 and 60.  The Gumbel temperature is gradually annealed from 5 to 1, for more stable training of this deep model. The mask unit is a squeeze unit with the same architecture as in SACT.
Results in Figure~\ref{fig:sota_imagenet} show that DynConv outperforms SACT by a large margin and matches the performances of the best layer-based methods. Those methods strongly benefit from the large number of residual blocks in ResNet-101 and therefore perform relatively better than they did in the CIFAR-10 experiment. 


\paragraph{Further analysis}

\begin{figure*}[t]
\begin{nscenter}
  \includegraphics[width=0.85\linewidth]{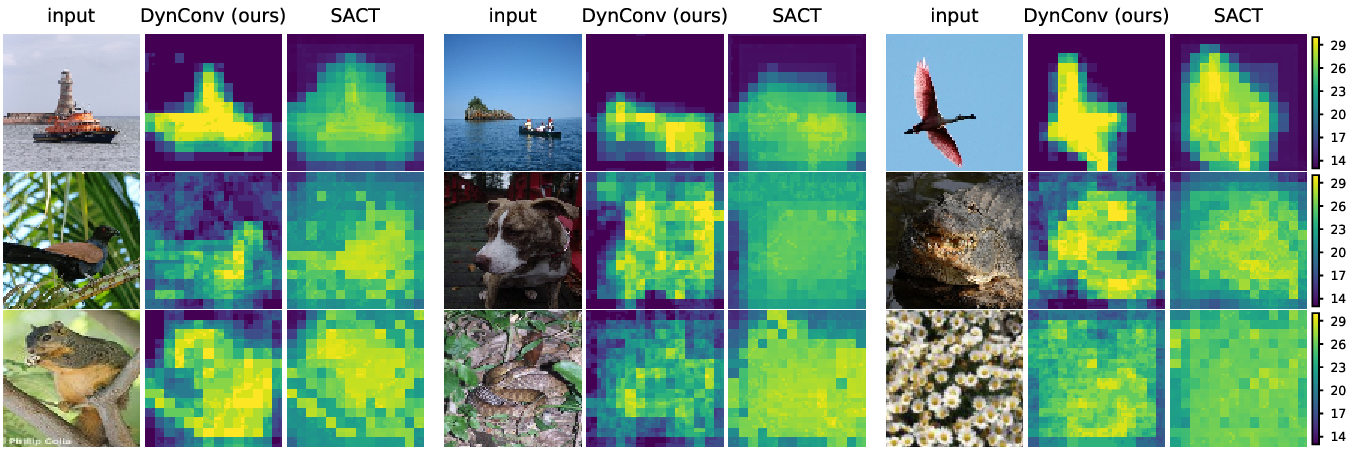}
  
\end{nscenter}
  \caption{Qualitative evaluation and comparison with SACT. Computational cost heatmaps indicate the number of residual blocks executed at each spatial location. Both methods have the same average complexity (5.7 GMACs), but ours shows better focus on the region of interest, both in simple examples (top row) as more cluttered ones (bottom row).  Computational cost heatmaps of SACT and input images obtained from~\cite{sact}. }
\label{fig:imagenet_qualitative}
\end{figure*}

\begin{figure}[tb]
\centering
\begin{subfigure}{.5\linewidth}
  \includegraphics[width=0.9\linewidth]{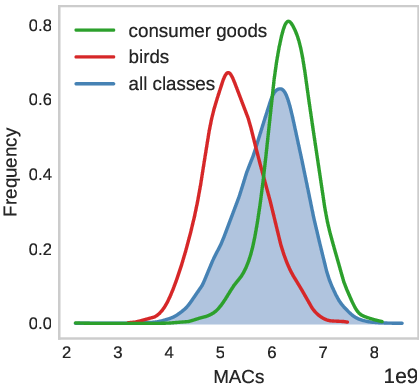}
    \caption{MACs per image}
     \label{fig:distributions_images}
\end{subfigure}%
\begin{subfigure}{.5\linewidth}
  \includegraphics[width=0.9\linewidth]{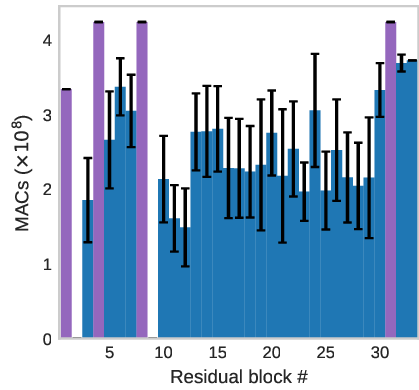}
   \caption{MACs per layer}
  \label{fig:distributions_layers}
\end{subfigure}
\caption{(a) Distribution of MACs over images in the ImageNet validation set. Images of birds, typically sparse, are processed with fewer operations than those of consumer goods. (b) Distribution of computations over residual blocks. Error bars indicate variance. Downsampling blocks (purple) are important and evaluated at all spatial positions.  }
\label{fig:distributions}
\end{figure}

Figure~\ref{fig:imagenet_qualitative} presents a qualitative comparison between our method and SACT. The number of computations per spatial location is visualized using \emph{computational cost heatmaps}, obtained by upscaling the binary execution masks of all residual blocks and subsequently summing them. Our method shows better focus on the region of interest. Analyzing the distribution of computations over ImageNet classes (Fig~\ref{fig:distributions_images}) shows that the network spends fewer computations on typically sparse images such as birds. 
When looking at the execution rates per residual block (Fig.~\ref{fig:distributions_layers}), it is clear that downsampling blocks are more important than others: all spatial locations are evaluated in these blocks. The last residual blocks, processing high-level features, are also crucial. This highlights the architectural advantage over SACT, where computation at a spatial location can only be halted. 

\subsubsection{DynConv inference speedup}

We integrate dynamic convolutions in MobileNetV2~\cite{mobilenetv2} and ShuffleNetV2~\cite{shufflenet}. Results for different computational budgets $\theta$ 
are shown in Table~\ref{tab:abl_mn2}. We use the Food-101 dataset~\cite{bossard2014food}, containing 75k/25k train/test images of 101 food classes, with InceptionV3~\cite{szegedy2016rethinking} data augmentation and image size $224{\times}224$. For MobileNetV2, reducing the computational budget $\theta$ results in a proportional increase of throughput (images per second). We compared a version with standard Gumbel-Softmax (Eq.~\ref{eq:gumbel}) to our reformulation (Eq.~\ref{eq:straight_through}): our reformulation (G-Binary) is significantly faster than the one with softmax and logarithms (G-Softmax).
We use ShuffleNetV2 with a residual connection~\cite{shufflenet}. Our dynamic convolutions are integrated in the convolutional branch of non-strided blocks. This architecture uses narrower residual blocks, and the relative cost of the mask unit with squeeze operation becomes significant. We suggest using a cheaper $1{\times}1$ convolution as mask unit for narrow networks. 




\begin{table}[tb]
\caption{Inference time of dynamic convolutions}
\label{tab:abl_mn2}
\small
\begin{tabularx}{\linewidth}{@{}lrrrr@{}}
\toprule
Method & Acc. & MACs & Im/Sec \\ \midrule
{MobileNetV2 x0.75~\cite{mobilenetv2} (our impl.)}    &  82.0    &    225M  &     508              \\
{MobileNetV2 x0.60 (our impl.)}    &  79.7    &    150M  &     638              \\
{$\theta=0.75$} &  81.2    &  200M    &  541                 \\
{$\theta=0.50$} &   80.6   & 174M     &    629               \\
{$\theta=0.25$ (G-Binary)} &  79.8    & 148M      & 724    \\ 
{$\theta=0.25$ (G-Softmax)} &  79.8    & 148M      & 522    \\ 
 \midrule
{ShuffleNetV2~\cite{shufflenet} (our impl.)}    & 78.7     &     149M &           710   \\
{$\theta=0.25$} & 76.5     &   100M   &      781             \\
{$\theta=0.25$ with $1{\times}1$ conv mask unit} &    76.3  & 97M     & 889                   \\
\bottomrule
\end{tabularx}
\end{table}

\subsection{Human pose estimation}

Human pose estimation is a task that is inherently spatially sparse: many pixels around the person are not relevant for keypoint detection and the output heatmaps are sparse. Most existing dynamic execution methods are not suited for this task: layer-based and channel-based methods, such as ConvNet-AIG~\cite{convnetaig}, do not exploit the spatial dimensions. SACT~\cite{sact} is not directly applicable on branched architectures such as stacked hourglass networks~\cite{newell2016stacked}, as it can only halt execution.

\paragraph{Experiment setup}
We base our work on Fast Pose Distillation (FPD)~\cite{zhang2019fast}, which uses the well-known stacked hourglass network for single-person human pose estimation~\cite{newell2016stacked}. Unlike their work, we do apply knowledge distillation since such method is complementary. The standard residual blocks are replaced by those of MobileNetV2 with depthwise convolution, in order to achieve efficient inference. Our baseline models have 96 features in the hourglass, expanded to 576 channels for the depthwise convolution. Models with different widths are obtained by multiplying the number of channels with a \emph{width multiplier} $\in \{0.5,0.75,1.0 \}$, while the network depth is varied by 
using 1, 2 and 4 hourglass stacks (1S, 2S and 4S). 
For dynamic convolutions, the computational budget of the base model is varied with $\theta \in \{0.125,0.25,0.5,0.75\}$. 

\begin{figure}[t]
\begin{nscenter}
  \includegraphics[width=0.75\linewidth]{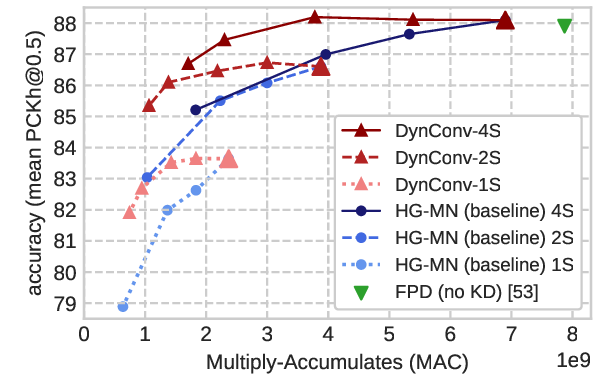}
\end{nscenter}
  \caption{Results on pose estimation (MPII validation set) for hourglass models with varying depths and widths. Our conditional execution method (DynConv, in red) always outperforms baseline models (in blue) with the same depth and number of computations. Our models achieve similar performance as FPD~\cite{zhang2019fast} (without knowledge distillation) with fewer computations.}
\label{fig:pose_quantitative}
\end{figure}

\begin{figure*}[tb]
\begin{nscenter}
  \includegraphics[width=0.75\linewidth]{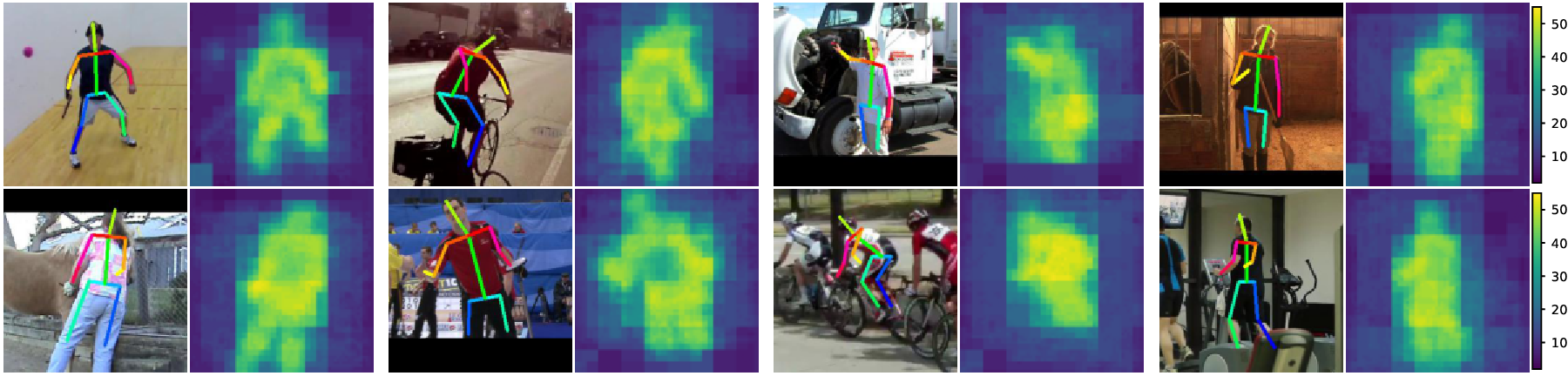}
 \end{nscenter}
  \caption{Computational cost on single-person human pose estimation (4-stack hourglass, $\theta = 0.125$). The network learns to apply the majority of convolutional operations on keypoint locations, even in the presence of clutter.
  }
\label{fig:pose_qualitative}
\end{figure*}

\begin{table*}[]
\caption{Timings on pose estimation for 2-stack models. Our models achieve significant wall-clock time speedup (measured in images processed per second) on an Nvidia GTX1050 Ti 4GB GPU. Timings of subcomponents (mask, bookkeeping, gather, residual function, scatter) are given as average time per execution, in milliseconds.   }
\small
\begin{tabularx}{\linewidth}{@{}lccccccccc@{}}
\toprule
                      Model   & PCKh@0.5 & \# Params                   & MACs & Images/Sec                & Mask & Bookkeeping               & Gather & Res. $\mathcal{F}$    & Scatter \\ \midrule
\multicolumn{1}{l|}{4-stack baseline}            & {88.1}& 6.88M & 6.90G & \multicolumn{1}{c|}{30} & N.A  & \multicolumn{1}{c|}{N.A.} & N.A    & 31.1              & N.A     \\
\multicolumn{1}{l|}{DynConv ($\theta = 0.75$)}  & {88.1}&  6.89M & 5.39G & \multicolumn{1}{c|}{33} & 0.48 & \multicolumn{1}{c|}{0.73} & 0.59   & 27.3              & 0.76    \\
\multicolumn{1}{l|}{DynConv ($\theta = 0.50$)}  & {88.2}&  6.89M & 3.78G & \multicolumn{1}{c|}{48} & 0.48 & \multicolumn{1}{c|}{0.78} & 0.47   & 18.4              & 0.54    \\
\multicolumn{1}{l|}{DynConv ($\theta = 0.25$)}  & {87.5}&  6.89M & 2.30G & \multicolumn{1}{c|}{67} & 0.45 & \multicolumn{1}{c|}{0.70} & 0.33   & 10.8               & 0.36    \\
\multicolumn{1}{l|}{DynConv ($\theta = 0.125$)} & {86.7}&  6.89M & 1.71G & \multicolumn{1}{c|}{85} & 0.50 & \multicolumn{1}{c|}{0.67} & 0.30   & 8.25               & 0.27   \\
\multicolumn{1}{l|}{baseline ($\times{0.5}$ channels)} & {85.2}&  1.83M & 1.83G & \multicolumn{1}{c|}{66} & N.A. & \multicolumn{1}{c|}{N.A.} & N.A.  &  14.1         & N.A.    \\ \midrule
\end{tabularx}
\label{table:pose_timings}
\end{table*}



\begin{table}[]
\small
\caption{Ablation on pose estimation (4-stack, $\theta = 0.125$).}
\begin{tabularx}{\linewidth}{@{}llcc@{}}
\toprule
Mask unit & Sparsity criterion & PCKh & Im/Sec \\ \midrule
$1{\times}1$ conv &  $\mathcal{L}_{sp,net}+\mathcal{L}_{sp,up}{+}\mathcal{L}_{sp,low}$ & {86.7} & 85 \\
squeeze unit & $\mathcal{L}_{sp,net}+\mathcal{L}_{sp,up}+\mathcal{L}_{sp,low}$ & 87.0 & 76\\
$1{\times}1$ conv &  $\mathcal{L}_{sp,net}$ (Eq.~\ref{eq:loss_net}) & {86.1}  & 85 \\
$1{\times}1$ conv &  $\mathcal{L}_{sp,per\_layer}$ (Eq.~\ref{eq:loss_layer}) & {86.3}  & 85\\
\bottomrule
\end{tabularx}
\label{table:pose_abl}
\end{table}

We use the MPII dataset~\cite{mpii} with standard test/validation split (22k/3k images). Images are resized to $256{\times}256$ and augmented with $\pm 30$ degrees rotation, $\pm 25$ percent scaling and random horizontal flip. 
No flip augmentation is used during evaluation. 
The optimizer is Adam with learning rate $2\text{e-}4$ for a batch size of 6 samples. The mean square error loss for heatmaps is averaged over samples. Training lasts for 100 epochs and the learning rate is reduced by factor 0.1 at epoch 60 and 90. The evaluation metric is the mean Percentage of Correct Keypoints, normalized by a fraction of the head size (PCKh@0.5), as implemented in~\cite{zhang2019fast}.

\paragraph{Results}
Figure~\ref{fig:pose_quantitative} shows that our method always outperforms non-conditional models with a similar number of operations. The number of operations is reduced by more than 45\% with almost no loss in accuracy. The performance difference between baselines of similar FLOPS and our method becomes larger for larger reductions of FLOPS. 

\begin{figure}[]
\centering
\begin{subfigure}{.465\linewidth}
  \centering
  \includegraphics[width=0.8\linewidth]{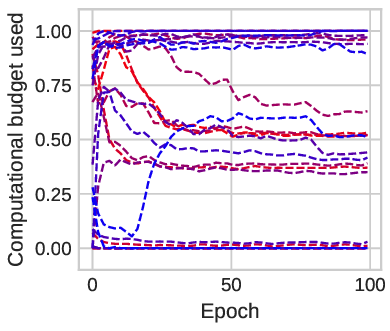}
  \caption{simple network criterion}
\end{subfigure}%
\begin{subfigure}{.535\linewidth}
  \centering
  \includegraphics[width=0.8\linewidth]{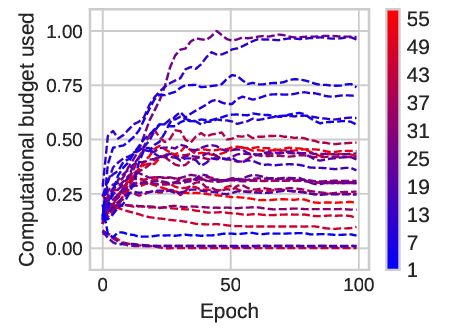}
  \caption{our annealing criterion}
\end{subfigure}
\caption{Evolution of the percentage of conditional computations executed, per residual block, during training of pose estimation. Early layers are colored blue, while deeper layers are red. The simple network-wise sparsity criterion  (Eq.~\ref{eq:loss_net}) causes early convergence to a suboptimal state, where many layers are not used. Our criterion where a lower and upper bound per block are annealed (Eq.~\ref{eq:finalloss}) provides better initialization and more stable training.}
\label{fig:annealing_criterion}
\end{figure}

Heatmaps in Figure~\ref{fig:pose_qualitative} show that the network learns to focus on the person.
Table~\ref{table:pose_timings} demonstrates that our DynConv method can significantly speed up inference, giving a 60\% speedup without loss in accuracy and a 125\% speedup with 0.6\% accuracy loss. The table also shows that the time spent on generating the mask, bookkeeping and gather/scatter operations is small compared to the cost of the residual function. Our model also outperforms smaller baseline model with equal FLOPS. \mbox{Table}~\ref{table:pose_abl} compares the squeeze masking unit used in SACT~\cite{sact} to a simple convolution. Using a squeeze function as mask unit performs slightly better at the cost of a lower inference speed. In addition, we compare the sparsity losses proposed in Section~\ref{sec:losses}. Our sparsity criterion with upper and lower bound outperforms more simple losses. The effect of our loss during training is shown in Figure~\ref{fig:annealing_criterion}. 




\section{Conclusion and future work}
In this work we proposed a method to speed up inference using dynamic convolutions. The network learns pixel-wise execution masks in an end-to-end fashion. Dynamic convolutions speed up inference and reduce the number of operations by only operating on these predicted locations. Our method achieves state-of-the-art results on image classification, and our qualitative analysis demonstrates the architectural advantages over existing methods. Our method is especially suitable for processing high-resolution images, \eg~in pose estimation or object detection tasks. 


\section{Acknowledgement}
This  work  was funded by IMEC through the ICON Lecture+ project and by CELSA through the SfS++ project.

 \clearpage

{\small
\bibliographystyle{ieee_fullname}
\bibliography{egbib}

\begin{thebibliography}{10}\itemsep=-1pt

\bibitem{dyanmiccapacity}
Amjad Almahairi, Nicolas Ballas, Tim Cooijmans, Yin Zheng, Hugo Larochelle, and
  Aaron Courville.
\newblock Dynamic capacity networks.
\newblock In {\em Proceedings of the International Conference on Machine
  Learning (ICML)}, pages 2549--2558, 2016.

\bibitem{mpii}
Mykhaylo Andriluka, Leonid Pishchulin, Peter Gehler, and Bernt Schiele.
\newblock 2d human pose estimation: New benchmark and state of the art
  analysis.
\newblock In {\em Proceedings of the IEEE Conference on Computer Vision and
  Pattern Recognition (CVPR)}, June 2014.

\bibitem{bejnordi}
Babak~Ehteshami Bejnordi, Tijmen Blankevoort, and Max Welling.
\newblock Batch-shaped channel gated networks.
\newblock {\em arXiv preprint arXiv:1907.06627}, 2019.

\bibitem{bengio2015conditional}
Emmanuel Bengio, Pierre-Luc Bacon, Joelle Pineau, and Doina Precup.
\newblock Conditional computation in neural networks for faster models.
\newblock {\em arXiv preprint arXiv:1511.06297}, 2015.

\bibitem{bengio2013deep}
Yoshua Bengio.
\newblock Deep learning of representations: Looking forward.
\newblock In {\em International Conference on Statistical Language and Speech
  Processing}, pages 1--37. Springer, 2013.

\bibitem{bengioST}
Yoshua Bengio, Nicholas L{\'e}onard, and Aaron Courville.
\newblock Estimating or propagating gradients through stochastic neurons for
  conditional computation.
\newblock {\em arXiv preprint arXiv:1308.3432}, 2013.

\bibitem{adaptivenetworks}
Tolga Bolukbasi, Joseph Wang, Ofer Dekel, and Venkatesh Saligrama.
\newblock Adaptive neural networks for efficient inference.
\newblock In {\em Proceedings of the 34th International Conference on Machine
  Learning (ICML) - Volume 70}, pages 527--536. JMLR.org, 2017.

\bibitem{bossard2014food}
Lukas Bossard, Matthieu Guillaumin, and Luc Van~Gool.
\newblock Food-101--mining discriminative components with random forests.
\newblock In {\em Proceedings of the European Conference on Computer Vision
  (ECCV)}, pages 446--461. Springer, 2014.

\bibitem{canziani2016analysis}
Alfredo Canziani, Adam Paszke, and Eugenio Culurciello.
\newblock An analysis of deep neural network models for practical applications.
\newblock {\em arXiv preprint arXiv:1605.07678}, 2016.

\bibitem{imagenet}
Jia Deng, Wei Dong, Richard Socher, Li-Jia Li, Kai Li, and Li Fei-Fei.
\newblock Imagenet: A large-scale hierarchical image database.
\newblock In {\em Proceedings of the IEEE Conference on Computer Vision and
  Pattern Recognition (CVPR)}, pages 248--255. IEEE, 2009.

\bibitem{droniou2015deep}
Alain Droniou, Serena Ivaldi, and Olivier Sigaud.
\newblock Deep unsupervised network for multimodal perception, representation
  and classification.
\newblock {\em Robotics and Autonomous Systems}, 71:83--98, 2015.

\bibitem{sact}
Michael Figurnov, Maxwell~D Collins, Yukun Zhu, Li Zhang, Jonathan Huang,
  Dmitry Vetrov, and Ruslan Salakhutdinov.
\newblock Spatially adaptive computation time for residual networks.
\newblock In {\em Proceedings of the IEEE Conference on Computer Vision and
  Pattern Recognition (CVPR)}, pages 1039--1048, 2017.

\bibitem{gao}
Xitong Gao, Yiren Zhao, Łukasz Dudziak, Robert Mullins, and Cheng zhong Xu.
\newblock Dynamic channel pruning: Feature boosting and suppression.
\newblock In {\em International Conference on Learning Representations}, 2019.

\bibitem{graham2014spatially}
Benjamin Graham.
\newblock Spatially-sparse convolutional neural networks.
\newblock {\em arXiv preprint arXiv:1409.6070}, 2014.

\bibitem{graham2017submanifold}
Benjamin Graham and Laurens van~der Maaten.
\newblock Submanifold sparse convolutional networks.
\newblock {\em arXiv preprint arXiv:1706.01307}, 2017.

\bibitem{graves}
Alex Graves.
\newblock Adaptive computation time for recurrent neural networks.
\newblock {\em arXiv preprint arXiv:1603.08983}, 2016.

\bibitem{han2015deep}
Song Han, Huizi Mao, and William~J. Dally.
\newblock Deep compression: Compressing deep neural network with pruning,
  trained quantization and huffman coding.
\newblock In {\em {ICLR}}, 2016.

\bibitem{resnet}
Kaiming He, Xiangyu Zhang, Shaoqing Ren, and Jian Sun.
\newblock Deep residual learning for image recognition.
\newblock In {\em Proceedings of the IEEE conference on Computer Vision and
  Pattern Recognition (CVPR)}, pages 770--778, 2016.

\bibitem{hinton2015distilling}
Geoffrey Hinton, Oriol Vinyals, and Jeff Dean.
\newblock Distilling the knowledge in a neural network.
\newblock {\em arXiv preprint arXiv:1503.02531}, 2015.

\bibitem{stochasticdepth}
Gao Huang, Yu Sun, Zhuang Liu, Daniel Sedra, and Kilian~Q Weinberger.
\newblock Deep networks with stochastic depth.
\newblock In {\em European Conference on Computer Vision (ECCV)}, pages
  646--661. Springer, 2016.

\bibitem{iandola2016squeezenet}
Forrest~N Iandola, Song Han, Matthew~W Moskewicz, Khalid Ashraf, William~J
  Dally, and Kurt Keutzer.
\newblock Squeezenet: Alexnet-level accuracy with 50x fewer parameters and 0.5
  mb model size.
\newblock {\em arXiv preprint arXiv:1602.07360}, 2016.

\bibitem{jacobs1991adaptive}
Robert~A Jacobs, Michael~I Jordan, Steven~J Nowlan, Geoffrey~E Hinton, et~al.
\newblock Adaptive mixtures of local experts.
\newblock {\em Neural computation}, 3(1):79--87, 1991.

\bibitem{gumbelsoftmax}
Eric Jang, Shixiang Gu, and Ben Poole.
\newblock Categorical reparameterization with gumbel-softmax.
\newblock In {\em Proceedings of the 5th International Conference on Learning
  Representations (ICLR)}, 2017.

\bibitem{kim2015compression}
Yong{-}Deok Kim, Eunhyeok Park, Sungjoo Yoo, Taelim Choi, Lu Yang, and Dongjun
  Shin.
\newblock Compression of deep convolutional neural networks for fast and low
  power mobile applications.
\newblock In {\em {ICLR} (Poster)}, 2016.

\bibitem{cifar}
Alex Krizhevsky, Geoffrey Hinton, et~al.
\newblock Learning multiple layers of features from tiny images.
\newblock Technical report, Citeseer, 2009.

\bibitem{lavin2016fast}
Andrew Lavin and Scott Gray.
\newblock Fast algorithms for convolutional neural networks.
\newblock In {\em Proceedings of the IEEE Conference on Computer Vision and
  Pattern Recognition (CVPR)}, pages 4013--4021, 2016.

\bibitem{lecun1990optimal}
Yann LeCun, John~S Denker, and Sara~A Solla.
\newblock Optimal brain damage.
\newblock In {\em Advances in Neural Information Processing Systems (NIPS)},
  pages 598--605, 1990.

\bibitem{li2016pruning}
Hao Li, Asim Kadav, Igor Durdanovic, Hanan Samet, and Hans~Peter Graf.
\newblock Pruning filters for efficient convnets.
\newblock In {\em International Conference on Learning Representations}, 2017.

\bibitem{li2017cascade}
Xiaoxiao Li, Ziwei Liu, Ping Luo, Chen Change~Loy, and Xiaoou Tang.
\newblock Not all pixels are equal: Difficulty-aware semantic segmentation via
  deep layer cascade.
\newblock In {\em Proceedings of the IEEE conference on Computer Vision and
  Pattern Recognition (CVPR)}, pages 3193--3202, 2017.

\bibitem{lin2017runtime}
Ji Lin, Yongming Rao, Jiwen Lu, and Jie Zhou.
\newblock Runtime neural pruning.
\newblock In {\em Advances in Neural Information Processing Systems (NIPS)},
  pages 2181--2191, 2017.

\bibitem{shufflenet}
Ningning Ma, Xiangyu Zhang, Hai-Tao Zheng, and Jian Sun.
\newblock Shufflenet v2: Practical guidelines for efficient cnn architecture
  design.
\newblock In {\em Proceedings of the European conference on computer vision
  (ECCV)}, pages 116--131, 2018.

\bibitem{maddison}
Chris~J. Maddison, Andriy Mnih, and Yee~Whye Teh.
\newblock The concrete distribution: {A} continuous relaxation of discrete
  random variables.
\newblock In {\em Proceedings of the 5th International Conference on Learning
  Representations (ICLR)}, 2017.

\bibitem{mnih2014recurrent}
Volodymyr Mnih, Nicolas Heess, Alex Graves, et~al.
\newblock Recurrent models of visual attention.
\newblock In {\em Advances in Neural Information Processing Systems (NIPS)},
  pages 2204--2212, 2014.

\bibitem{molchanov2016pruning}
Pavlo Molchanov, Stephen Tyree, Tero Karras, Timo Aila, and Jan Kautz.
\newblock Pruning convolutional neural networks for resource efficient
  inference.
\newblock In {\em International Conference on Learning Representations}, 2017.

\bibitem{newell2016stacked}
Alejandro Newell, Kaiyu Yang, and Jia Deng.
\newblock Stacked hourglass networks for human pose estimation.
\newblock In {\em Proceedings of the European conference on Computer Vision
  (ECCV)}, pages 483--499. Springer, 2016.

\bibitem{panda2016conditional}
Priyadarshini Panda, Abhronil Sengupta, and Kaushik Roy.
\newblock Conditional deep learning for energy-efficient and enhanced pattern
  recognition.
\newblock In {\em Design, Automation \& Test in Europe Conference \& Exhibition
  (DATE)}, pages 475--480. IEEE, 2016.

\bibitem{sbnet}
Mengye Ren, Andrei Pokrovsky, Bin Yang, and Raquel Urtasun.
\newblock Sbnet: Sparse blocks network for fast inference.
\newblock In {\em Proceedings of the IEEE Conference on Computer Vision and
  Pattern Recognition (CVPR)}, pages 8711--8720, 2018.

\bibitem{romero2014fitnets}
Adriana Romero, Nicolas Ballas, Samira~Ebrahimi Kahou, Antoine Chassang, Carlo
  Gatta, and Yoshua Bengio.
\newblock Fitnets: Hints for thin deep nets.
\newblock In {\em ICLR (Poster)}, 2015.

\bibitem{sainath2013low}
Tara~N Sainath, Brian Kingsbury, Vikas Sindhwani, Ebru Arisoy, and Bhuvana
  Ramabhadran.
\newblock Low-rank matrix factorization for deep neural network training with
  high-dimensional output targets.
\newblock In {\em IEEE International Conference on Acoustics, Speech and Signal
  Processing (ICASSP)}, pages 6655--6659. IEEE, 2013.

\bibitem{mobilenetv2}
Mark Sandler, Andrew Howard, Menglong Zhu, Andrey Zhmoginov, and Liang-Chieh
  Chen.
\newblock Mobilenetv2: Inverted residuals and linear bottlenecks.
\newblock In {\em Proceedings of the IEEE Conference on Computer Vision and
  Pattern Recognition (CVPR)}, pages 4510--4520, 2018.

\bibitem{sharma2015action}
Shikhar Sharma, Ryan Kiros, and Ruslan Salakhutdinov.
\newblock Action recognition using visual attention.
\newblock In {\em Neural Information Processing Systems (NIPS) Time Series
  Workshop}, December 2015.

\bibitem{sindhwani2015structured}
Vikas Sindhwani, Tara Sainath, and Sanjiv Kumar.
\newblock Structured transforms for small-footprint deep learning.
\newblock In {\em Advances in Neural Information Processing Systems (NIPS)},
  pages 3088--3096, 2015.

\bibitem{sze2017efficient}
Vivienne Sze, Yu-Hsin Chen, Tien-Ju Yang, and Joel~S Emer.
\newblock Efficient processing of deep neural networks: A tutorial and survey.
\newblock {\em Proceedings of the IEEE}, 105(12):2295--2329, 2017.

\bibitem{szegedy2016rethinking}
Christian Szegedy, Vincent Vanhoucke, Sergey Ioffe, Jon Shlens, and Zbigniew
  Wojna.
\newblock Rethinking the inception architecture for computer vision.
\newblock In {\em Proceedings of the IEEE Conference on Computer Vision and
  Pattern Recognition (CVPR)}, pages 2818--2826, 2016.

\bibitem{mnasnet}
Mingxing Tan, Bo Chen, Ruoming Pang, Vijay Vasudevan, Mark Sandler, Andrew
  Howard, and Quoc~V Le.
\newblock Mnasnet: Platform-aware neural architecture search for mobile.
\newblock In {\em Proceedings of the IEEE Conference on Computer Vision and
  Pattern Recognition (CVPR)}, pages 2820--2828, 2019.

\bibitem{branchynet}
Surat Teerapittayanon, Bradley McDanel, and Hsiang-Tsung Kung.
\newblock Branchynet: Fast inference via early exiting from deep neural
  networks.
\newblock In {\em 23rd International Conference on Pattern Recognition (ICPR)},
  pages 2464--2469. IEEE, 2016.

\bibitem{convnetaig}
Andreas Veit and Serge Belongie.
\newblock Convolutional networks with adaptive inference graphs.
\newblock In {\em Proceedings of the European Conference on Computer Vision
  (ECCV)}, pages 3--18, 2018.

\bibitem{veit2016residual}
Andreas Veit, Michael~J Wilber, and Serge Belongie.
\newblock Residual networks behave like ensembles of relatively shallow
  networks.
\newblock In {\em Advances in Neural Information Processing Systems (NIPS)},
  pages 550--558, 2016.

\bibitem{skipnet}
Xin Wang, Fisher Yu, Zi-Yi Dou, Trevor Darrell, and Joseph~E Gonzalez.
\newblock Skipnet: Learning dynamic routing in convolutional networks.
\newblock In {\em Proceedings of the European Conference on Computer Vision
  (ECCV)}, pages 409--424, 2018.

\bibitem{wen2016learning}
Wei Wen, Chunpeng Wu, Yandan Wang, Yiran Chen, and Hai Li.
\newblock Learning structured sparsity in deep neural networks.
\newblock In {\em Advances in Neural Information Processing Systems (NIPS)},
  pages 2074--2082, 2016.

\bibitem{wu2016quantized}
Jiaxiang Wu, Cong Leng, Yuhang Wang, Qinghao Hu, and Jian Cheng.
\newblock Quantized convolutional neural networks for mobile devices.
\newblock In {\em Proceedings of the IEEE Conference on Computer Vision and
  Pattern Recognition (CVPR)}, pages 4820--4828, 2016.

\bibitem{blockdrop}
Zuxuan Wu, Tushar Nagarajan, Abhishek Kumar, Steven Rennie, Larry~S Davis,
  Kristen Grauman, and Rogerio Feris.
\newblock Blockdrop: Dynamic inference paths in residual networks.
\newblock In {\em Proceedings of the IEEE Conference on Computer Vision and
  Pattern Recognition (CVPR)}, pages 8817--8826, 2018.

\bibitem{zhang2019fast}
Feng Zhang, Xiatian Zhu, and Mao Ye.
\newblock Fast human pose estimation.
\newblock In {\em Proceedings of the IEEE Conference on Computer Vision and
  Pattern Recognition}, pages 3517--3526, 2019.

\end{thebibliography}
}

\end{document}